\documentclass[10pt,twocolumn,letterpaper]{article}

\usepackage[pagenumbers]{cvpr} 
\usepackage{booktabs}
\usepackage{multirow}
\usepackage{algorithm}
\usepackage{algpseudocode}
\usepackage{amsmath}
\usepackage{pifont}
\usepackage{xcolor}
\usepackage{xspace}
\usepackage{rotating}
\usepackage{array} 
\usepackage[table]{xcolor}

\definecolor{cvprblue}{HTML}{0071BC}
\definecolor{mygreen}{rgb}{0.13, 0.55, 0.13}
\definecolor{myrowcolor}{HTML}{E6F2FA}
\definecolor{mygray}{gray}{0.6}
\definecolor{namecolor}{RGB}{168,119,200}
\definecolor{de-emphasized-color}{gray}{.6}
\definecolor{deepblue}{RGB}{31,78,121}
\definecolor{deepyellow}{RGB}{127,96,0}

\usepackage[bookmarks=False,pagebackref,breaklinks,colorlinks,citecolor=cvprblue]{hyperref}
\newcommand{\no}{\textcolor{Red}{\ding{55}}}
\newcommand{\yes}{\textcolor{mygreen}{\ding{51}}}
\newcommand{\ours}{\textcolor{namecolor}{\textbf{LAST}}\xspace}

\def\paperID{} 
\def\confName{CVPR}
\def\confYear{2026}

\title{\ours: \textcolor{namecolor}{L}e\textcolor{namecolor}{A}rning to Think in \textcolor{namecolor}{S}pace and \textcolor{namecolor}{T}ime \\
for Generalist Vision-Language Models}
\author{
Shuai Wang\textsuperscript{1} \quad
Daoan Zhang\textsuperscript{2} \quad
Tianyi Bai\textsuperscript{3} \quad
Shitong Shao\textsuperscript{1}\quad
Jiebo Luo\textsuperscript{2} \quad
Jiaheng Wei\textsuperscript{1} \vspace{.2em} \\
\textsuperscript{1}HKUST(GZ) \quad  \textsuperscript{2} University of Rochester\quad
\textsuperscript{3}HKUST \\
}

\begin{document}
\maketitle
\begin{abstract}
Humans can perceive and understand 3D space and long videos from sequential visual observations. But do vision-language models (VLMs) can? Recent work demonstrates that even state-of-the-art VLMs still struggle to understand 3D space and long videos, although they are powerful in typical vision-language tasks. Current methods often rely on specialized architectural designs to improve performance for 3D tasks and video understanding tasks \textbf{separately}. In contrast, we propose \ours, short for \textcolor{namecolor}{L}e\textcolor{namecolor}{A}rn to Think in \textcolor{namecolor}{S}pace and \textcolor{namecolor}{T}ime, to \textbf{jointly} improve 3D spatial and long video understanding for general VLMs with only a set of 2D images as inputs. \ours makes VLMs think in space and time rather than only with text before giving the final answer, building visual thinking trajectories in 3D space and temporal dimension. We demonstrate the effectiveness of \ours in two scenarios: 1) zero-shot, where we directly prompt proprietary models; and 2) fine-tuning general VLMs with data that include thinking trajectories in 3D space and time. We show that \ours brings substantial gains in various benchmarks, including 3 spatial understanding, 4 video understanding, and 3 image understanding tasks. Notably, \textbf{15.8}\% gains on EgoSchema with GPT-4o in a zero-shot manner and \textbf{8.3}\% gains on VSI-Bench compared with Qwen2.5-VL-7B.

\vspace{-1.6em}
\end{abstract}

\section{Introduction}
\label{sec:intro}
Understanding spatial and temporal information from multiple visual observations is important for general visual intelligence. Building from powerful large language models (LLMs)~\cite{gpt3,llama2,instructgpt}, vision-language models (VLMs) have shown remarkable capacities on typical visual-language tasks~\cite{gpt4o,comanici2025gemini2.5,llava,bai2025qwen25vl,cambrian1,zhu2025internvl3}. Despite strong performance on visual-language tasks, many recent works~\cite{majumdar2024openeqa,vsibench} show that even state-of-the-art VLMs still struggle with 3D spatial~\cite{vsibench,liu2025coarse} and long video understanding tasks~\cite{lvbench}.

Prior works focus on separately enhancing spatial and temporal understanding ability. To improve the 3D spatial understanding of VLMs, many attempts have been made, including two main directions: 1) using 3D data as inputs for VLMs~\cite{chen2024spatialvlm,cai2025spatialbot,SpatialRGPT,3dllm,chen2024ll3da,deng20253dllava,video3dllm}; and 2) introducing or designing specialized architectures for 3D spatial understanding tasks~\cite{3dllm,wu2025spatialmllm}. Similarly, to enable VLMs to understand long videos, previous works have focused on specific model designs~\cite{papalampidi2024simple,mc-vit} or token compression~\cite{song2024moviechat}. Although previous work has proposed some effective methods and made progress, solving the two problems above separately hinders the development of a unified and general-purpose VLMs. In this paper, we study the following question.

\textit{Can general-purpose VLMs be enhanced to reason spatially and temporally without relying on additional inputs or task-specific architectures?}

To address this question, we first revisit how VLMs perform reasoning for complex problems. Most VLMs achieve strong reasoning ability with \textit{language-centric} reasoning with chain-of-thought (CoT)~\cite{cot}. CoT generates more \textit{text} tokens to ``think'' before giving the final answer. However, this paradigm fails in spatial and temporal understanding (see~\cite{vsibench} and~\cref{tab:text_video_cot}). The reason behind this is that text-only CoT ignores the rich and continuous visual world and lacks visual engagement, which is important for solving complex geometry problems~\cite{visual_sketchpad}, visual search~\cite{v_star}, and spatial reasoning~\cite{visual_sketchpad,vsibench}. To tackle the problem, we aim to inject visual information, including 3D spatial and temporal information into the thinking process, \ie to make VLMs ``think in space and time''.

\begin{figure*}[t]
    \centering
    \includegraphics[width=0.99\linewidth]{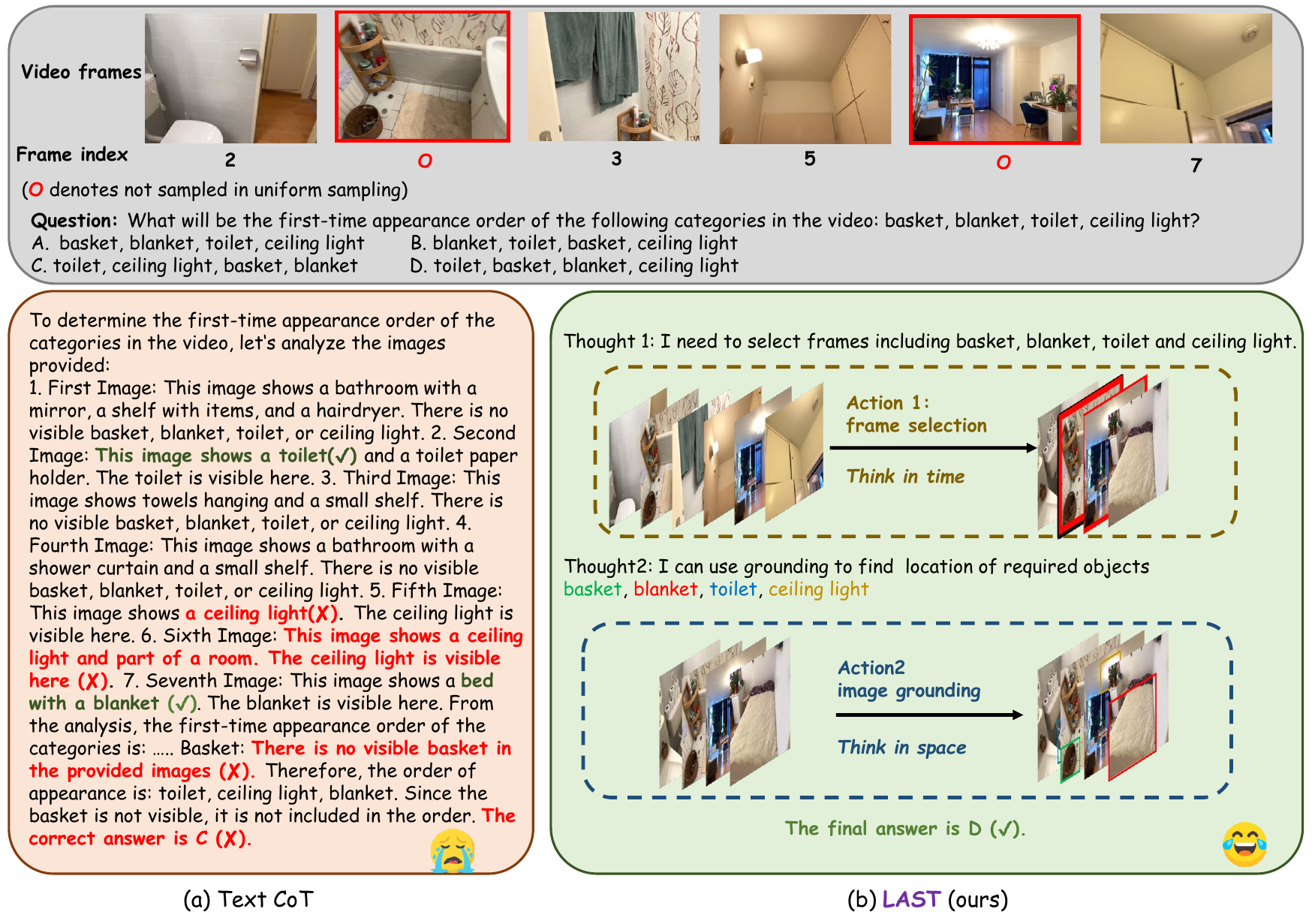}
    \vspace{-1em}
    \caption{\textbf{Comparison of text CoT and \ours}. (a) CoT for VLMs suffers from fixed visual context and generates wrong reasoning traces. CoT fails to capture important frames (\ie cannot see the basket due to missing frames between the second and third sampled frames) and generates hallucinations in the 5th frame (wall light, not ceiling light). However, in (b), \ours can \textcolor{deepyellow}{\textbf{think in time}} (use frame selection tools to re-sample video frames and newly sampled video frames are marked with \textcolor{red}{red} in Action 1) and \textcolor{deepblue}{\textbf{think in space}} (use grounding tools to identify objects). \ours achieves the correct solution by building intermediate visual trajectories.}
    \label{fig:teaser}    
\vspace{-1.5em}
\end{figure*}

Our key idea is to explicitly incorporate visual tokens into the reasoning process, forming visual chains of thought. However, most current VLMs take textual and visual inputs but can only generate text tokens, which prevents them from naturally producing visual reasoning steps. To solve this, we use external tools to generate visual tokens, such as object tracking~\cite{sam2} and grounding~\cite{grounding_dino}, to build visual tokens as demonstrated in~\cref{fig:teaser} and~\cref{fig:example5}. We name our method \ours, short for \textcolor{namecolor}{L}e\textcolor{namecolor}{A}rn to Think in \textcolor{namecolor}{S}pace and \textcolor{namecolor}{T}ime. We study \ours in two ways: zero-shot and fine-tuning. In a zero-shot manner, we prompt powerful but close-weight models, such as GPT-4o~\cite{gpt4o}. On the other hand, to make our approach applicable to open-weight and lightweight models, we further construct a dataset annotated with spatial and temporal reasoning traces, including 48K samples with text-based CoT and 124K samples with visual thinking trajectories, and fine-tune Qwen2.5-VL-7B~\cite{bai2025qwen25vl} to obtain \ours-7B.

\vspace{-.8em}
We extensively evaluate \ours on three spatial reasoning and four video understanding benchmarks. In the zero-shot setting, where \ours is applied to GPT‑4o~\cite{gpt4o}, our method consistently improves performance across all seven benchmarks. For instance, on spatial reasoning tasks, \ours yields \textbf{9.6}\% accuracy gain on SQA3D~\cite{sqa3d} without relying on any 3D inputs. On video understanding tasks, it achieves \textbf{15.8}\% improvement on EgoSchema~\cite{egoschema} (69.6$\rightarrow$85.4) with only eight input frames. For the fine-tuned model \ours‑7B, performance improvements remain consistent across all benchmarks. Specifically, \ours‑7B attains 41.3\% accuracy on VSI‑Bench~\cite{vsibench}, outperforming Qwen2.5‑VL‑7B~\cite{bai2025qwen25vl} by \textbf{8.3}\%. On NExT‑QA~\cite{nextqa}, it achieves 86.2\% on the validation set and 78.0\% on the challenging ATP‑hard subset. These results collectively demonstrate that introducing visual chains of thought substantially enhances both spatial and temporal understanding, validating the effectiveness and generality of our proposed approach \ours.

To summarize, we make the following contributions: (1) We propose \ours, a unified approach for general VLMs to improve spatial and temporal understanding ability without specific architecture design and extra inputs. (2) We build large-scale data with visual thinking trajectories and deploy \ours-7B that significantly improves the base model. (3) \ours and \ours-7B are validated through extensive evaluations on spatial and video understanding benchmarks, demonstrating superior effectiveness.
\section{Related Work}
\label{sec:related_work}
\noindent\textbf{Spatial understanding}. Based on powerful LLMs~\cite{gpt3,touvron2023llama,llama2}, VLMs exhibit visual understanding ability for 2D images~\cite{gpt4o,team2024gemini,bai2025qwen25vl,lin2024vila,llava,li2025llavaonevision,flamingo,comanici2025gemini2.5}. However, 3D spatial understanding, relevant to robotics~\cite{ZitkovichYXXXXW23,palme,song2025robospatial} and autonomous driving~\cite{tian2024drivevlm}, presents significant challenges. Some efforts~\cite{cai2025spatialbot,SpatialRGPT,3dllm,chen2024ll3da,deng20253dllava,video3dllm,wu2025spatialmllm,zheng2025learning,cheng20253d} improve 3D spatial understanding by using additional 3D inputs, such as point clouds, depth images, and camera features or specific modules and designs (like VGGT encoder~\cite{wang2025vggt}). In contrast, our approach only requires videos as input and uses general VLMs without any task-specific designs to explore \textit{general} visual spatial intelligence.

\noindent\textbf{Video understanding}. While VLMs demonstrate impressive performance on standard video understanding tasks~\cite{bai2025qwen25vl,glm45v,li2025llavaonevision,song2024moviechat,li2024mvbench,apollo}, understanding high-frame-rate and long-duration videos is challenging due to high computational and memory demands. Furthermore, VLMs fail to capture useful information in long-dependency tasks~\cite{hong2025motionbench,lvbench,LongVideoBench}. Previous works explore methods of video compression to tackle this, including adopting Q-Former~\cite{blip2,internvideo2} for video feature extraction and fusing neighboring frames~\cite{qwen2vl}. Different from the above, we explore making VLMs ``think'' in space and time for videos, which is applicable for general VLMs without any specific model design.

\noindent\textbf{Thinking beyond text}. LLMs/VLMs achieve strong reasoning ability with \textit{language-centric} reasoning, \ie CoT~\cite{cot,selfconsistency}. By decomposing complex problems into a sequence of textual reasoning steps, CoT has significantly enhanced VLMs in visual reasoning questions~\cite{internvl25,wang2025athena,DBLP:journals/tmlr/0001Z00KS24}. While simple and effective, text-only CoT provides limited exploration of intelligence and ignores the rich and continuous visual world. Furthermore, many tasks require visual engagement, such as solving complex geometry problems~\cite{visual_sketchpad}, visual search~\cite{v_star}, and spatial reasoning~\cite{visual_sketchpad,vsibench}. To tackle this, ``thinking with images'' is proposed to inject visual information into the thinking process~\cite{su2025thinkingwithimage,openai_thinking_with_images}. Since most VLMs cannot generate visual tokens, previous work mainly uses external tools~\cite{vis_program,qi2025cogcom,menon2025caviar,wu2023visualchatgpt,bai2025multi,vis_program_distill}, such as code, visual expert models, zoom-in-out, and so on. Previous work mainly focuses on solving tasks with 2D images, lacking exploration in spatial and temporal understanding. In this work, we shift the research line from 2D images to 3D space and videos.

\section{Method}
\label{sec:method}
\subsection{Formulation}
Given a question $Q$ and an initial input sequence of visual observations $\mathcal{I}_0=\{I_i\}_{i=1}^T$ in an environment with $T$ images, a VLM $\mathcal{M}$ will give the text response sequence
\begin{equation}
    \mathcal{A}=\mathcal{M}\left(Q,\mathcal{I}_0\right).
    \label{eq:vlm_infer}
\end{equation}
Note that theses observations do not need to be images from the video, and they can also be a set of images depicting the 3D scene, \ie from multiple viewpoints.

To improve spatial and temporal understanding ability of VLMs, a simple way is to use CoT~\cite{cot} for VLMs. CoT improves reasoning ability of VLMs by generating more text tokens to ``think'' before giving the final answer. However, text-only CoT fails for understanding 3D space or temporal information~\cite{vsibench}. We propose to enhance VLMs with \textit{thinking in space and time} by constructing a chain of vision thoughts $\mathcal{V}$ in 3D space and temporal dimensions, and we could get the final answer with the help of vision chains as
\begin{equation}
    \mathcal{A}=\mathcal{M}\left(Q,\mathcal{I}_0,\mathcal{V}\right).
\end{equation}
Due to VLMs usually generate text tokens rather than vision tokens, we consider using external tools for input visual observations to build intermediate visual chains $\mathcal{V}$ during inference stage. Tools we consider are introduced as follows.

\begin{figure}[t]
    \centering
    \includegraphics[width=0.95\linewidth]{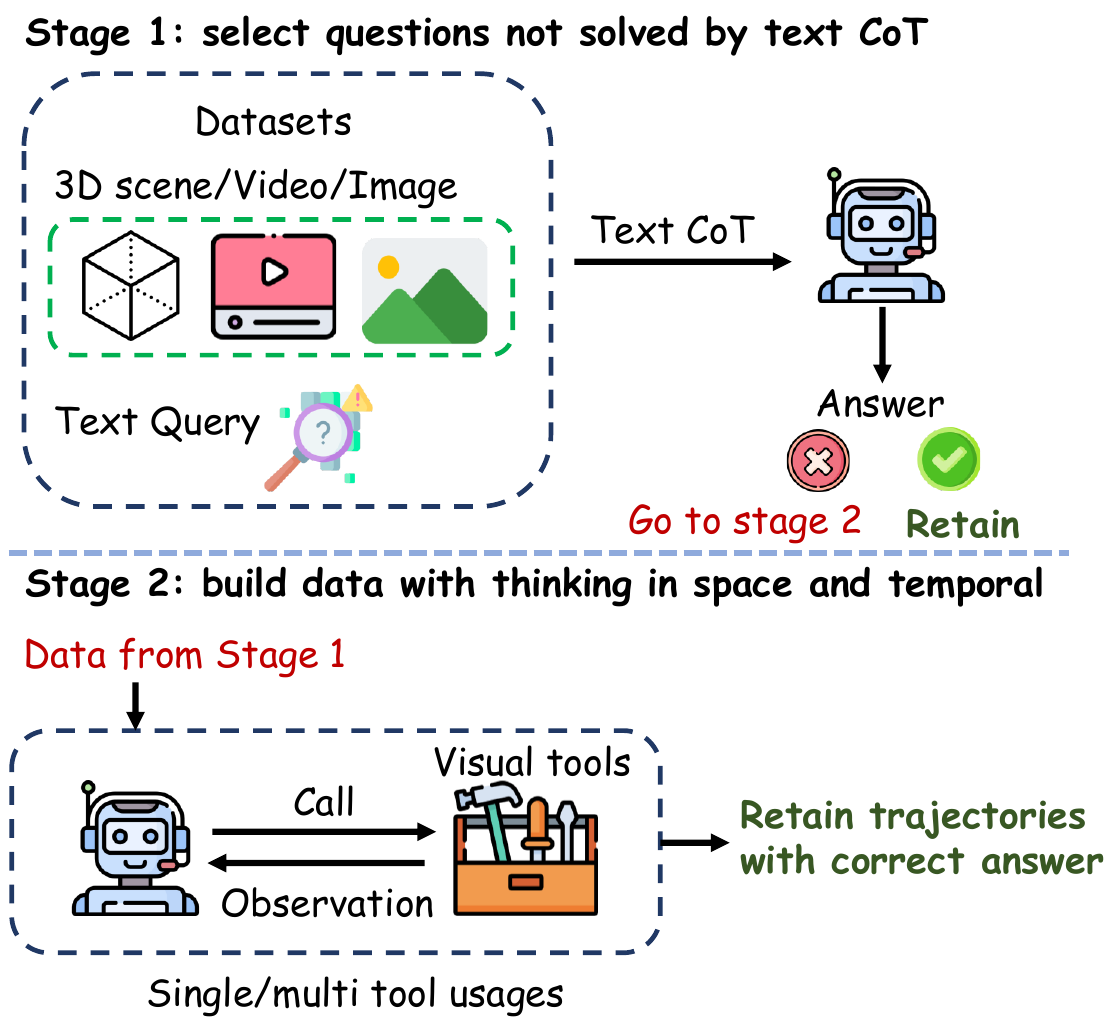}
    \vspace{-0.1in}
    \caption{\textbf{The illustration of data curation pipeline}. In the Stage 1, we prompt VLMs with text CoT and only retain sample with the correct answer. In the Stage 2, we prompt VLMs to use external visual tools to solve questions that can not solved by text CoT. Finally we collect data with text thinking trajectories in Stage 1 and visual thinking trajectories in Stage 2.}
    \label{fig:data_pipeline}
\vspace{-0.2in}
\end{figure}

\subsection{Tools}
\label{sec:method_tools}
We aim for a minimal but sufficient set of tools for spatial and video understanding. Tools we consider include: frame selection, object tracking, and temporal grounding. In addition, we also add useful tools for image understanding, including depth estimation, zoom-in-out, and image grounding. We provide details of each tool we use as follows.

\noindent\textbf{Frame selection}. Given a text query $Q$ and a sequence of visual observations in an environment $\mathcal{I}_0$, the standard inference approach for VLMs to answer a question $Q$ about an input $\mathcal{I}_0$ is to forward it through the model $\mathcal{M}$ as shown in~\cref{eq:vlm_infer}. Although current VLMs are now capable of handling increasingly large input context lengths~\cite{liu2023ring,team2024gemini}, they often fail to leverage long context effectively and are confused by irrelevant context~\cite{language_repo_video_und,liu2023lost}. Therefore, selecting an appropriate number of relevant frames is important. We denote this stage as
\begin{equation}
    \begin{aligned}
        \mathcal{S} = G\left(\mathcal{I}_0, Q\right), \quad
        \mathcal{A} = \mathcal{M}\left(Q, \mathcal{S}\right),
    \end{aligned}
\end{equation}
where $\mathcal{S}=\{I_{s_{i}}\}_{i=1}^{K}$ is the set of selected images. $K$ is the number of selected images and $s_k \in \{1,2,\ldots,T\}$ denotes the index of selected images.

The choice of $G$ is flexible, and we only consider training-free methods for frame selection for simplicity and efficiency. A straightforward solution is to select frames via uniform sampling, \ie, we uniformly sample $K$ frames from a sequence of visual observations $\mathcal{I}_0$. However, uniform sampling ignores the relation between the text query $Q$ and the input frames, which may sample frames that are irrelevant to the query or ignore frames that contain information relevant to the query. Therefore, some query-aware image selection methods are proposed to select query-related frames~\cite{q_frame,AKS}. However, only selecting related frames may introduce visually redundant frames, which means some highly similar frames are chosen. Furthermore, we propose to simultaneously maximize the relevance between the text query and the selected images and minimize the visual redundancy for frame selection.

\textbf{Maximize text-image relevance.}~~  For solving problems with multiple frames, the relevance between each frame and the prompt is important. A simple method is to choose the top-k similar frames with query $Q$, \ie
\begin{equation}
    \mathcal{S} = \textrm{top-k}\{k|q^{\textrm{T}}i_k\},
    \label{eq:relevance_topk}
\end{equation}
where $q \in \mathbb{R}^{d}$ and $i_k \in \mathbb{R}^d$ denote $\ell_2$-normalized embedding of query $Q$ and $I_k$, respectively. We use SigLIP-2~\cite{tschannen2025siglip2} to extract the text and image embedding and select the most similar images with the text query $Q$. 

\begin{table*}[t]
\centering
\caption{\textbf{Evaluation results on ScanQA~\cite{azuma2022scanqa} and SQA3D~\cite{sqa3d}}. We report the results on validation set of ScanQA and test set of SQA3D following previous work~\cite{wu2025spatialmllm,video3dllm}. For general models with video inputs only, we sample 16 frames per video.}
\vspace{-0.1in}
\resizebox{\textwidth}{!}{
    \begin{tabular}{l*{8}{c}}
        \toprule
         \multirow{2}{*}{\textbf{Method}} & \multirow{2}{*}{\textbf{Video input only}} & \multicolumn{5}{c}{\textbf{ScanQA}} & \multicolumn{2}{c}{\textbf{SQA3D}} \\
        \cmidrule(lr){3-7}\cmidrule(lr){8-9}
         & & BLEU-1 & BLEU-4 & METEOR & ROUGE-L & CIDEr & EM-1 & EM-R1 \\
        \midrule
        \multicolumn{1}{l}{\textit{Task-specific models}} & & & & & & & & \\
        ScanQA~\cite{azuma2022scanqa}  & \no & 30.2 & 10.1 & 13.1 & 33.3 & 64.9 & 47.2 & --  \\
        SQA3D~\cite{sqa3d}  & \no & 30.5 & 11.2 & 13.5 & 34.5 & -- & 46.6 & -- \\
        3D-Vista~\cite{3dvista}  & \no & -- & -- & 13.9 & 35.7 & -- & 48.5 & --\\
        \midrule
        \multicolumn{1}{l}{\textit{VLMs with 3D inputs or specialized architectures}} & & & & & & & & \\
        3D-LLM~\cite{3dllm}  & \no & 39.3 & 12.0 & 14.5 & 35.7 & 69.4 & -- & -- \\
        LL3DA~\cite{chen2024ll3da}   & \no & -- & 13.5 & 15.9 & 37.3 & 76.8 & -- & -- \\
        Chat-Scene~\cite{chat_scene} & \no & 43.2 & 14.3 & 18.0 & 41.6 & 87.7 & 54.6 & 57.5 \\
        3D-LLaVA~\cite{deng20253dllava}  & \no & -- & 17.1 & 18.4 & 43.1 & 92.6 & 54.5 & 56.6 \\
        Video-3D LLM~\cite{video3dllm} & \no & 47.1 & 16.2 & 19.8 & 49.0 & 102.1 & 58.6 & -- \\
        LLaVA-3D~\cite{LLaVA-3D} & \no & -- & 16.4 & 20.8 & 49.6 & 103.1 & 60.1 & -- \\
        Spatial-MLLM-4B~\cite{wu2025spatialmllm} & \yes & 44.4 & 14.8 & 18.4 & 45.0 & 91.8 & 55.9 & 58.7 \\
        \midrule
        \multicolumn{1}{l}{\textit{Proprietary models}} & & & & & & & & \\
        GPT-4o~\cite{gpt4o} & \yes & 31.1 & 13.0 & 12.0 & 34.2 & 65.4 & 47.8 &	49.6\\ 
        \rowcolor{myrowcolor}
        + \ours & \yes & 38.4  &  14.7 &  16.7 & 44.5 & 76.8 & 58.4 &  61.2\\
        \midrule
        \multicolumn{1}{l}{\textit{Open-weight models}} & & & & & & & & \\
         Qwen2.5-VL-3B~\cite{bai2025qwen25vl} & \yes & 22.5 & 3.8 & 9.7 & 25.4 & 47.4 & 43.4 & 45.9 \\         
         Qwen2.5-VL-72B~\cite{bai2025qwen25vl} & \yes & 26.8 & 12.0 & 13.0 & 35.2 & 66.9 & 47.0 & 50.9 \\
         Oryx-34B~\cite{oryxmllm}  & \yes & 38.0 & -- & 15.0 & 37.3 & 72.3 & -- & -- \\
         LLaVA-Video-7B~\cite{zhang2025llavavideo} & \yes & 39.7 & 3.1 & 17.7 & 44.6 & 88.7 & 48.5 & -- \\         
         Qwen2.5-VL-7B~\cite{bai2025qwen25vl} & \yes & 27.8 & 3.0 & 11.4 & 29.3 & 53.9 & 46.5 & 49.8 \\
         \rowcolor{myrowcolor}
         \ours-7B & \yes & 41.2 & 10.4 & 19.1 &  41.8 & 66.4 & 56.1 & 59.6 \\
        \bottomrule
    \end{tabular}
}
\vspace{-0.1in}
\label{tab:scanqa_sqa3d}
\end{table*}
\begin{table*}[t]
\centering
\caption{\textbf{Evaluation results on VSI-Bench~\cite{vsibench}}. Spatial-MLLM-4B~\cite{wu2025spatialmllm} is \textcolor{de-emphasized-color}{de-emphasized} because it uses VGGT~\cite{wang2025vggt} as the encoder.}
\vspace{-0.1in}
\resizebox{\textwidth}{!}
{
    \begin{tabular}{lcccccccccc}
        \toprule
         \multirow{2}{*}{\textbf{Method}} & \multirow{2}{*}{\textbf{Frame}} & \multicolumn{4}{c}{\textbf{Numerical Answer}} & \multicolumn{4}{c}{\textbf{Multiple-Choice Answer}} & \multirow{2}{*}{\textbf{Avg.}} \\
        \cmidrule(lr){3-6}\cmidrule(lr){7-10}
         & & Obj. Cnt. & Abs. Dist. & Obj. Size & Room Size & Rel. Dist. & Rel. Dir. & Route Plan & Appr. Order & \\
        \midrule
        \textit{Proprietary models} \\        
        Gemini-1.5 Pro~\cite{team2024gemini} & 1 fps & 56.2 & 30.9 & 64.1 & 43.6 & 51.3 & 46.3 & 36.0 & 34.6 & 45.4 \\
        GPT-4o~\cite{gpt4o} & 16 & 41.4 & 22.8 & 55.4 & 54.7 & 44.4 & 40.4 & 33.0 & 35.3 & 40.9\\
        \rowcolor{myrowcolor}
        + \ours & 16 & 39.2 & 28.3 & 61.2 & 46.7 & 53.7 & 41.5 & 42.8 & 59.2 & 46.6\\
        \midrule
        \textit{Open-weight models} \\
        \textcolor{de-emphasized-color}{Spatial-MLLM-4B}~\cite{wu2025spatialmllm} & \textcolor{de-emphasized-color}{16} & \textcolor{de-emphasized-color}{65.3} & \textcolor{de-emphasized-color}{34.8} & \textcolor{de-emphasized-color}{63.1} & \textcolor{de-emphasized-color}{45.1} & \textcolor{de-emphasized-color}{41.3} & \textcolor{de-emphasized-color}{46.2} & \textcolor{de-emphasized-color}{33.5} & \textcolor{de-emphasized-color}{46.3} & \textcolor{de-emphasized-color}{48.4} \\
        Qwen2.5VL-3B~\cite{bai2025qwen25vl} &16  & 24.3 & 24.7 & 31.7 & 22.6 & 38.3 & 41.6 & 26.3 & 21.2 & 30.6 \\
        Qwen2.5VL-72B~\cite{bai2025qwen25vl} & 16& 25.1 & 29.3 & 54.5 & 38.8 & 38.2 & 37.0 & 34.0 & 28.9 & 37.0 \\
        VILA-1.5-40B~\cite{lin2024vila} & 32 & 22.4  & 24.8 & 48.7 & 22.7 & 40.5 & 25.7 & 31.5 & 32.9 & 31.2 \\
        LLaVA-OneVision-72B~\cite{li2025llavaonevision}& 32  & 43.5 & 23.9 & 57.6 & 37.5 & 42.5 & 39.9 & 32.5 & 44.6 & 40.2 \\
        LLaVA-Video-72B~\cite{zhang2025llavavideo} & 32 & 48.9 & 22.8 & 57.4 & 35.3 & 42.4 & 36.7 & 35.0 & 48.6 & 40.9 \\  
        LongVILA-8B~\cite{chen2025longvila} & 32 & 29.1 & 9.1 & 16.7 & 0.0 & 29.6 & 30.7 & 32.5 & 25.5 & 21.6 \\
        LongVA-7B~\cite{zhang2025llavavideo} & 32& 38.0 & 16.6 & 38.9 & 22.2 & 33.1 & 43.3 & 25.4 & 15.7 & 29.2 \\              
        Video-R1-7B~\cite{feng2025videor1} & 64  & -- & -- & -- & -- & -- & -- & -- & --  & 37.1 \\
        STAR-R1-7B~\cite{zheng2025starr} & 64 & -- & -- & -- & -- & -- & -- & -- & --  & 34.1 \\               
        Qwen2.5VL-7B~\cite{bai2025qwen25vl} & 16 & 40.9 & 14.8 & 43.4 & 10.7 & 38.6 & 38.5 & 33.0 & 29.8 & 33.0 \\         
        \rowcolor{myrowcolor}
        \ours-7B & 16 & 51.1& 28.8 & 48.4 & 31.8 & 43.5 & 46.3 & 34.0 & 46.4 & 41.3 \\
        \bottomrule
    \end{tabular}
}
\vspace{-0.2in}
\label{tab:vsibench}
\end{table*}

However, only considering the most relevant frames introduces too many redundant and similar images. We propose minimizing \textbf{the visual redundancy} of selected images and maximizing the diversity of selected images. The process can be modeled as a determinantal point process~\cite{odile_process,fast_dpp}. Formally, for $T$ frames, we aim to select $K$ frames to maximize the diversity of selected frames. We first get the image embedding $i_k$ of each image $I_k$ and compute the image similarity matrix $L_S\in \mathbb{R}^{K \times K}$, where $L_{p,q}=\exp \left(i_p^{\top} i_q\right)$ denotes the similarity between $i_p$ and $i_q$. The objective function is
\begin{equation}
     \mathcal{S} = \operatorname*{argmax}  \det\left(L_S\right), \quad \textrm{s.t.} \quad \lvert\mathcal{S}\rvert=K,
\label{eq:dpp}
\end{equation}
where $\det(L_S)$ denotes the determinant of matrix $L_S$. The geometric meaning of~\cref{eq:dpp} is obvious: it represents selecting $K$ vectors from $T$ vectors such that the volume spanned by these $K$ vectors is maximized. If the volume spanned by the selected $K$ frames of images is the largest, it indicates that their diversity is optimal. Because solving~\cref{eq:dpp} is NP-hard, we use the fast greedy algorithm introduced in~\cite{fast_dpp}. See~\cref{sec:appendix_algo} for more details.

We combine~\cref{eq:relevance_topk} and~\cref{eq:dpp} to select the most important images. First, we choose the most relevant images with~\cref{eq:relevance_topk}. The number of selected images is set to $4K$. After that, we use~\cref{eq:dpp} to choose $K$ images. We compare our method with some recently proposed training-free methods, such as Q-Frame~\cite{q_frame} and AKS~\cite{AKS}. The results are shown in~\cref{tab:comp_frame_select}, which shows that our method achieves the best performance on EgoSchema~\cite{egoschema}. See~\cref{sec:appenidx_results} in the Appendix for more experimental results.

\noindent\textbf{Object tracking}. Understanding object correspondence relationships (for 3D space)  and object movement (for videos) during multiple images is important. However, current VLMs do not support object tracking for multiple visual observations. Here we use SAM2~\cite{sam2} (SAM2.1-Hiera-Large) as the tracking tool. For visual observations $\mathcal{I}$ and the target object $c$, we use SAM2 to identify and follow the movement of the object $c$ in a visual sequence. Formally, given input visual observations $\mathcal{I}$ and the target object $c$, we get the tracking results $\mathcal{I}^{\prime}=\textrm{SAM2}\left(\mathcal{I},c\right)$, where $\mathcal{I}^{\prime}$ denotes visual observations with marks to highlight the movement of the object $c$. For each object correspondence relationship, we visualize the correspondences directly in the image as a marker. Specifically, for each object, we overlay a circle with a radius of 4 pixels at the object center onto the image. We also try to use the whole mask of the object. However, simply adding marks achieves the best performance. We analyze that using the whole mask occludes visual content, leading to poor results.

\noindent\textbf{Temporal grounding}. It is important to know when something starts and ends for video understanding. Temporal grounding allows VLMs to use temporal information by trimming out the relevant segment. This can also enable better interpretability in knowing which part of the video the agent chose to consider. We apply TFVTG~\cite{tfvtg} as the temporal grounding tool. For visual observations $\mathcal{I}$ and the text query $Q$, temporal grounding aims to localize relevant video segments that correspond specifically to given textual queries, enabling detailed interaction with video
content. Temporal grounding also reduces the visual tokens and improves efficiency.

\noindent\textbf{Image grounding}. For a single image, grounding aims to localize a specific region within an image based on a given textual description. Although current VLMs show strong grounding ability, we choose Grounding-DINO-L~\cite{grounding_dino} with 341 M parameters for efficiency.

\noindent\textbf{Depth estimation}. Depth estimation provides relative depth for different objects, which is helpful for many 3D spatial understanding tasks. For depth estimation, we use Depth-Anything(-Large)~\cite{depth_anything} with 335 M parameters for image and video depth estimation.

\noindent\textbf{Zoon-in-out}. We use OpenCV~\cite{opencv} to zoom in on a specified area of an image.

\subsection{Enhancing VLMs with \ours}
\label{sec:method_distil}
It is straightforward to prompt proprietary models such as GPT-4o~\cite{gpt4o} to simulate thinking in space and time in a zero-shot manner. To study whether we could integrate \ours into small open-weight models and the generalization of \ours, we collect data from public sources and fine-tune Qwen2.5-VL-7B~\cite{bai2025qwen25vl} to get \ours-7B.

\noindent\textbf{Data curation}. To build high-quality and diverse training data, we design a two-stage data curation pipeline, as shown in~\cref{fig:data_pipeline}. First, we collect public datasets including images, videos, and 3D scenes, and corresponding questions. For 3D scene data, we only use the video parts and do not use 3D features as inputs, unlike~\cite{deng20253dllava,3dllm}. For all video inputs, we first downsample videos at 4 fps and constrain the number of pixels in each frame to be less than 50,176. Given a question about an image or video, we first determine whether this question could be solved by text-only CoT in the first stage. If the question could be solved by text CoT, we retain text thinking trajectories. For questions not solved by text CoT, we introduce external tools in~\cref{sec:method_tools} to obtain thinking trajectories in space and time in the second stage. Finally, we get training samples with text-thinking trajectories and visual thinking trajectories. In total, we collected 48K samples with text-based CoT and 124K samples with visual thinking trajectories, averaging 1.72 tool calls per sample. More details are provided in~\cref{sec:appendix_data}.

\noindent\textbf{Training details}. We choose Qwen2.5-VL-7B~\cite{bai2025qwen25vl} as the base model, and fine-tune Qwen2.5-VL-7B using SWIFT~\cite{zhao2025swift} with standard next token prediction loss. The global batch size is set to 32, and the number of epochs is set to one. We use AdamW~\cite{adamw} with the cosine learning rate schedule and the learning rate is set to 1e-6.

\section{Experiments}
\subsection{Setup}
\textbf{Benchmarks and metrics}. We first evaluate \ours on three spatial understanding tasks. \textbf{ScanQA}~\cite{azuma2022scanqa} and \textbf{SQA3D}~\cite{sqa3d} are two 3D question-answering benchmarks. For ScanQA~\cite{azuma2022scanqa}, we report BLEU-1/4~\cite{papineni2002bleu}, METEOR~\cite{banerjee2005meteor}, ROUGE-L~\cite{lin2004rouge}, CIDEr~\cite{vedantam2015cider} for the validation set that includes 4675 questions from 71 scenes, following previous work~\cite{wu2025spatialmllm,3dllm,video3dllm}. For SQA3D~\cite{sqa3d}, we evaluate on its test set that contains 3519 QA pairs. Because SQA3D provides clear answers, we use exact match accuracy (EM-1) and relaxed accuracy (EM-R1) as evaluation metrics. \textbf{VSI-Bench}~\cite{vsibench} contains 5130 questions collected from egocentric videos sourced from ScanNet~\cite{scannet}, ScanNet++~\cite{scannet_pp}, and ARKitScenes~\cite{arkit_scenes}. VSI-Bench~\cite{vsibench} includes two task types: Multiple-Choice Answer (MCA) and Numerical Answer (NA). We report accuracy for MCA tasks and relative accuracy across confidence threshold $\mathcal{C}=\{0.5,0.55,\cdots,0.95\}$ for NA tasks following~\cite{vsibench}.

Furthermore, we consider four video understanding benchmarks. \textbf{EgoSchema}~\cite{egoschema} consists of 3-minute-long clips, each with a question and 5 answer choices. We use the validation set of 500 questions that have publicly available answers. \textbf{NExT-QA}~\cite{nextqa} is a video question answering benchmark. We use its validation set with 4996 multiple-choice questions for our evaluation. Additionally, following previous work~\cite{sevila,li2024mvbench}, we also report results on the more challenging ATP-hard subset. \textbf{LVBench}~\cite{lvbench} includes 1549 multiple-choice questions with an average video length of over one hour, which measures models’ ability to process and reason about extended temporal sequences. \textbf{MotionBench}~\cite{hong2025motionbench} evaluates fine-grained motion comprehension in video-understanding models. We report results on the dev set, which contains 4018 multiple-choice questions. For the above four video benchmarks, we report accuracy because they only include multiple-choice questions.

Although our work focuses on 3D spatial and temporal understanding, for a complete evaluation, we also add three image benchmarks: HR-Bench~\cite{hr_bench} (including 4K and 8K subsets), V*~\cite{v_star}, and BLINK~\cite{BLINK} that are widely used in previous works~\cite{bai2025multi,visual_sketchpad}. 

\noindent\textbf{Implementation details}. For zero-shot experiments, we use GPT-4o~\cite{gpt4o} as the base model. During inference, we usually sample 16 frames for all benchmarks except 32 frames for LVBench~\cite{lvbench}. For fair comparison, when \ours performs frame selection or temporal grounding, \ours selects the same number of frames as the baseline.

\noindent\textbf{Baselines}.~ For ScanQA and SQA3D, we compare with specialized models~\cite{azuma2022scanqa,sqa3d,3dvista}, 3D VLMs with 3D inputs or specific architecture~\cite{3dllm,video3dllm} and general-purpose VLMs. For VSI-Bench and other video understanding tasks, we compare \ours with model architectures specifically for long video understanding~\cite{llovi,papalampidi2024simple}, general VLMs~\cite{bai2025qwen25vl,glm45v} and agentic methods~\cite{fan2024videoagent,wang2024videoagent,zhi2025videoagent2}.

\begin{table}[t]
\centering
\caption{\textbf{Results on EgoSchema~\cite{egoschema} validation set}.}
\vspace{-0.1in}
    \begin{tabular}{lcc}
    \toprule
    \textbf{Method} & \textbf{Frame} & \textbf{EgoSchema}\\
    \midrule
    LongViViT~\cite{papalampidi2024simple} & 256 & 56.8 \\
    MC-ViT-L~\cite{mc-vit} & 128 & 62.6 \\
    LLoVi~\cite{llovi} & 180 & 58.3 \\
    VideoAgent~\cite{wang2024videoagent} & -- & 60.2 \\    
    VideoAgent~\cite{fan2024videoagent} & -- & 62.8 \\
    LangRepo~\cite{language_repo_video_und} & -- & 66.2 \\    
    VideoAgent2~\cite{zhi2025videoagent2} & -- & 80.6 \\
    LifelongMemory~\cite{wang2023lifelongmemory} & -- & 72.0 \\
    Gemini-1.5-flash~\cite{team2024gemini} & 1 fps & 72.9 \\
    TCoT~\cite{arnab2025temporal} & 1 fps & 75.2 \\
    MVU~\cite{ranasinghe2024understanding} & 16 & 60.3 \\
    VideoChat2~\cite{li2024mvbench} & 16 & 63.6 \\
    \midrule 
    GPT-4o~\cite{gpt4o} & 8 & 69.6 \\
    \rowcolor{myrowcolor}
    + \ours & 8 & 85.4\\
    \midrule
    Qwen2.5-VL-7B~\cite{bai2025qwen25vl} & 16 & 55.0 \\
    \rowcolor{myrowcolor}
    \ours-7B & 16 & 69.0\\
    \bottomrule
    \end{tabular}
\label{tab:egoschema}
\vspace{-0.2in}
\end{table}
\begin{table}[t]
    \centering
    \caption{\textbf{Results on NExT-QA~\cite{nextqa}}.}
    \vspace{-0.1in}
    \begin{tabular}{lccc}
        \toprule
         \multirow{2}{*}{\textbf{Method}}  & \multirow{2}{*}{\textbf{Frame}} & \multicolumn{2}{c}{\textbf{NExT-QA}}\\
        \cmidrule(lr){3-4}
           & & Val & ATP-hard\\
        \midrule
        VideoAgent~\cite{wang2024videoagent} & 8.2 & 71.3 & 58.4 \\
        VideoAgent2~\cite{zhi2025videoagent2} & -- & 80.5 & 68.2 \\
        SeViLA~\cite{sevila} & -- & 63.6 & 50.8 \\        
        VFC~\cite{vfc} &32 & 51.5 & 31.4 \\
        ViperGPT~\cite{suris2023vipergpt} & -- & 60.0 & -- \\
        LLoVi~\cite{llovi} & 1 fps & 73.8 & -- \\
        TCoT~\cite{arnab2025temporal} & 1 fps & 81.0 & -- \\
        VideoChat2~\cite{li2024mvbench}& 16  & 79.5 & 68.2 \\
        \midrule
        GPT-4o~\cite{gpt4o} &16&  78.9 & 72.5 \\
        \rowcolor{myrowcolor}
        + \ours & 16& 84.8 & 76.6\\
        \midrule
        Qwen2.5-VL-7B~\cite{bai2025qwen25vl} & 16 &79.6 & 71.8 \\
        \rowcolor{myrowcolor}
        \ours-7B & 16 & 86.2 & 78.0 \\
        \bottomrule
    \end{tabular}
\vspace{-0.1in}
\label{tab:nextqa}
\end{table}
\begin{table}[t]
\centering
\caption{\textbf{Results on LVBench~\cite{lvbench} and MotionBench~\cite{hong2025motionbench}}. $\dagger$: 32 frames for LVBench and 16 frames for MotionBench.}
\vspace{-0.1in}
\resizebox{\linewidth}{!}{
    \begin{tabular}{lccc}
    \toprule
    \textbf{Method} & \textbf{Frame} & \textbf{LVBench} & \textbf{MotionBench} \\
    \midrule
    GLM-4.5V~\cite{glm45v} & 2 fps & 56.2 & 61.4 \\
    GLM-4.1V-9B~\cite{glm45v} & 2 fps & 44.0 & 58.7 \\
    Gemini-1.5-Pro~\cite{team2024gemini} & 1 fps & 33.1 & 51.0 \\
    PLLaVA-34B~\cite{xu2024pllava} & 16 & 26.1 & 52.0 \\
    \midrule
    GPT-4o~\cite{gpt4o} & 16 & 32.9 & 53.5  \\
    \rowcolor{myrowcolor}
    + \ours & 16 & 40.6 & 56.2\\
    \midrule
    Qwen2.5-VL-7B~\cite{bai2025qwen25vl} & 32/16$^\dagger$ & 39.2 & 58.6\\
    \rowcolor{myrowcolor}
    \ours-7B & 32/16$^\dagger$ & 43.8 & 62.0 \\
    \bottomrule
    \end{tabular}
}
\vspace{-0.1in}
\label{tab:lvbench_motion}
\end{table}
\begin{table}[t]
\centering
\caption{Results of \textbf{text CoT \vs \ours}. \colorbox{green!15}{\small Green} denotes improved performance and \colorbox{red!15}{\small red} denotes worse.}
\vspace{-0.1in}
\resizebox{\linewidth}{!}{
    \begin{tabular}{lcccc}
    \toprule
    \multirow{2}{*}{\textbf{Method}} & \multirow{2}{*}{\textbf{VSI-Bench}} & \multirow{2}{*}{\textbf{EgoSchema}} & \multicolumn{2}{c}{\textbf{NExT-QA}}\\
        \cmidrule(lr){4-5}
        &   & & Val & ATP-hard\\
    \midrule
    GPT-4o~\cite{gpt4o} & 40.9 & 69.6 & 78.9 & 72.5 \\
    + CoT~\cite{cot} & \cellcolor{green!15}42.5 & \cellcolor{red!15}62.7 & \cellcolor{red!15}77.5 & \cellcolor{red!15}72.3 \\
    + \ours & \cellcolor{green!15}46.6 & \cellcolor{green!15}85.4 & \cellcolor{green!15}84.8 & \cellcolor{green!15}76.6 \\
    \bottomrule
    \end{tabular}
}
\vspace{-0.2in}
\label{tab:text_video_cot}
\end{table}

\subsection{Main Results}
\noindent\textbf{Zero-shot results.} \Cref{tab:scanqa_sqa3d} shows results on ScanQA~\cite{azuma2022scanqa} and SQA3D~\cite{sqa3d}. It is shown that \ours generally improves GPT-4o on both datasets. For example, GPT-4o with \ours improves 11.6 points on EM-R1 on SQA3D and 11.4 points on CIDEr on ScanQA \textbf{with only video inputs}. Furthermore, \ours achieves comparable performance with 3D-specific models~\cite{video3dllm}, \eg, 58.4 \textit{vs.} 58.6 on SQA3D. These results demonstrate the potential of \ours to improve 3D scene understanding with general VLMs in a zero-shot manner. Furthermore, we present the quantitative results on VSI-Bench~\cite{vsibench} in \cref{tab:vsibench}. First, by applying \ours, we significantly improve GPT-4o, achieving the best performance with 16 frames. By thinking in spatial-temporal, general VLMs such as GPT-4o can achieve comparable performance with specifically-designed models like Spatial-MLLM~\cite{wu2025spatialmllm}. Results on three spatial understanding benchmarks show that our method can significantly improve the spatial understanding of general VLMs.

\cref{tab:egoschema}-\ref{tab:lvbench_motion} show results on four video understanding benchmarks. \ours generally improve GPT-4o across four benchmarks, \eg 15.8 points on EgoSchema~\cite{egoschema} (\cref{tab:egoschema}) and 7.7 points on LVBench~\cite{lvbench} (\cref{tab:lvbench_motion}), in a zero-shot manner. Notably, \ours achieves better performance with fewer input frames. For example, with only 8 frames, our method gets 85.4\% on the validation set of EgoSchema~\cite{egoschema}, while VideoAgent2~\cite{zhi2025videoagent2} gets 60.2\% and TCoT~\cite{arnab2025temporal} gets 75.2\% with 1 fps (almost 180 frames per video for EgoSchema~\cite{egoschema}).

\noindent\textbf{Results on \ours-7B}. \cref{tab:scanqa_sqa3d}-\ref{tab:lvbench_motion} show results on 7 benchmarks of \ours-7B. Built from Qwen2.5-VL-7B~\cite{bai2025qwen25vl}, \ours-7B significantly improves Qwen2.5-VL-7B on \textit{all} benchmarks. For example, \ours-7B achieves 69.0\% accuracy on EgoSchma~\cite{egoschema} with 16 frames and 14 points gains compared with Qwen2.5-VL-7B~\cite{bai2025qwen25vl} in~\cref{tab:egoschema}. \ours-7B gets 86.2\% on the validation set of NExT-QA~\cite{nextqa} and 78.0\% on the more challenging ATP-hard subset, which outperforms all baselines.

Results on zero-shot experiments and \ours-7B show that \ours can be adopted in general VLMs in different ways and significantly improves the spatial and temporal understanding abilities of general VLMs. In addition, we present results of three image benchmarks in~\cref{sec:appenidx_results} and show that \ours is also applicable for 2D images.

\subsection{Analysis}
\noindent\textbf{Thinking in space and time \vs text thinking}.  To illustrate the benefits of thinking in space and temporal over thinking with only text, we compare it with text-only CoT on three benchmarks. Results shown in~\cref{tab:text_video_cot} indicate that text-only CoT~\cite{cot} degrades performance on some benchmarks, \eg, EgoSchema~\cite{egoschema} and NExT-QA~\cite{nextqa}. However, \ours, thinking in space and time, achieves consistent and significant improvement across three benchmarks. Furthermore, \ours achieves better performance compared with text-only CoT, \eg, 85.4 \vs 62.7 on EgoSchema~\cite{egoschema}, and 84.8 \vs 77.5 on the validation set of NExT-QA~\cite{nextqa}. In addition, we study the effectiveness of different visual context length and show that \ours stills outperforms text CoT in various visual context length, see~\cref{sec:appenidx_results} for more details.
\vspace{-0.1in}
\begin{figure*}[t]
    \centering
    \includegraphics[width=\linewidth]{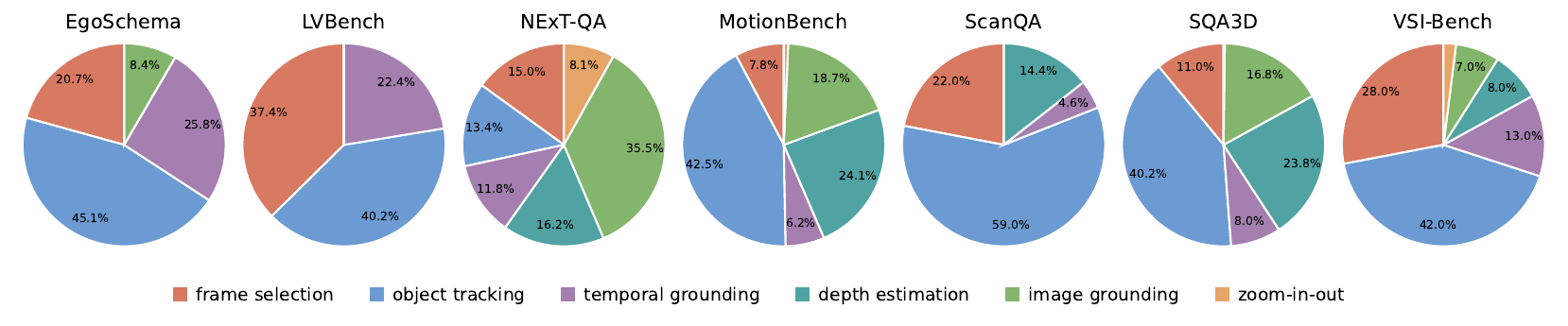}
    \vspace{-0.35in}
    \caption{\textbf{Percentage of tools} GPT-4o uses for different benchmarks.}
    \label{fig:tool_usage}
\vspace{-0.15in}
\end{figure*}
\begin{table}
    \centering
    \caption{\textbf{Ablation on external tools} with GPT-4o.}
    \vspace{-0.1in}
        \resizebox{\linewidth}{!}{
            \begin{tabular}{cccccccc}
            \toprule
             FS  & OT & TG & IG & DE & ZI & VSI-Bench &  EgoSchema\\
            \midrule
            & & & & & & 40.9 & 69.6\\
             &\yes & & & & & 43.8 & 76.8 \\
             \yes & & & & & & 41.4 & 73.2 \\
             & & \yes & & & & 41.6 & 72.6 \\
             \yes & \yes & \yes & & & & 45.8 & 83.6 \\
             \yes& \yes & \yes & \yes & \yes & \yes & 46.6 & 85.4\\
            \bottomrule
            \end{tabular}
        }
    \flushleft
    \vspace{-0.1in}
    \begin{flushleft}
    \footnotesize
    \textbf{Abbreviation.} FS: frame selection, OT: object tracking, TG: temporal grounding, IG: image grounding, DE: depth estimation, ZI: zoom-in-out.
    \end{flushleft}
\vspace{-0.15in}
\label{tab:ablation_tool}
\end{table}
\begin{figure*}[t]
    \centering
    \includegraphics[width=0.95\textwidth]{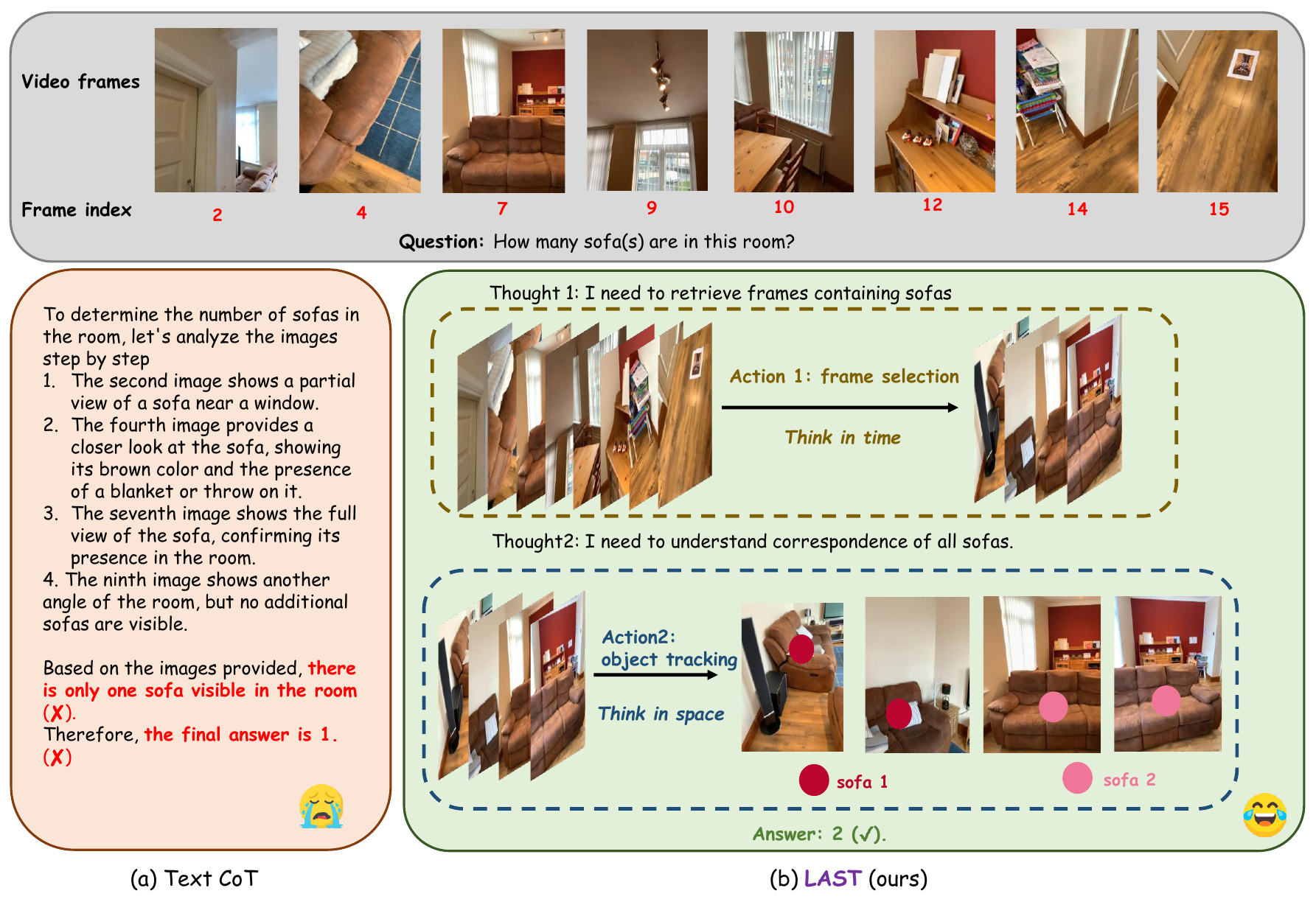}
    \vspace{-0.1in}
    \caption{Qualitative comparison of text-only CoT and \ours on VSI-Bench~\cite{vsibench} with GPT-4o. We highlight errors with \textcolor{red}{red}. GPT-4o with text CoT fails to understand correspondence of different sofas appeared in the video. In contrast, \ours could understand correspondence of two sofas in the video and get the correct solution. For clarity, we enlarge marks in frames.}
    \label{fig:example5}
\vspace{-0.2in}
\end{figure*}

\noindent\textbf{Analysis on tool usage.} To understand how \ours uses different external tools to solve problems, we analyze the distribution of tool usage for different tasks in~\cref{fig:tool_usage}. For long video understanding tasks such as LVBench~\cite{lvbench}, object tracking, temporal grounding, and frame selection play an important role. With the help of temporal grounding and frame selection, VLMs focus on relevant visual context and ignore other frames. Object tracking helps establish correspondence in long videos, which helps VLMs capture information about object movement. For 3D spatial understanding tasks, object tracking and depth estimation matter. The reason behind this is that object tracking can capture object movement in long videos and multi-view of 3D scene, which helps 3D scene understanding. ~\cref{tab:ablation_tool} shows detailed results, and the results are the same.

\noindent\textbf{Thinking pattern analysis}.  To understand the difference between text CoT and \ours, we present a qualitative example in~\cref{fig:example5}. For the question \textit{How many sofa(s) are in this room?}, VLMs need to be able to retrieve frames that include sofas and understand correspondence of them, \ie ``Are the sofas appearing in different frames of the video the same one?''. VLMs with text CoT fail to understand correspondence, although two sofas have appeared in the video. However, equipped with \ours, VLMs can use external tools to re-sample video frames (to retrieve all frames including sofas) and identify the object correspondence in every frame with the extra object tracking model.
\vspace{-.5em}
\section{Conclusion}
\vspace{-.2em}
We propose \ours, a method that makes VLMs think in spatial and temporal dimensions, thereby improving spatial and temporal understanding. By using external tools, \ours builds thinking chains in 3D space and time rather than in text alone. \ours not only works on proprietary models like GPT-4o but can also be adapted to open-weight models through fine-tuning. Experimental results show that \ours achieves significant gains on various benchmarks. Furthermore, we analyze the thinking process of \ours and text-only CoT, demonstrating the superiority of thinking in spatial and temporal dimensions over text-based thinking.
\clearpage
{
    \small
    \bibliographystyle{ieeenat_fullname}
    \bibliography{main}
}

\clearpage
\appendix
\section{Training Data Details}
\label{sec:appendix_data}
We collect videos and images from multiple public data sources, as shown in~\cref{tab:data}. For videos, we use all tools introduced~\cref{sec:method_tools}. For images, we only use depth estimation, image grounding, zoom-in-out. Totally, we collected 48K samples with text CoT and 124K samples with visual thinking trajectories, averaging 1.72 tool calls per sample. We validate the effectiveness of each part in~\cref{sec:appenidx_results}.
\section{Minimize Visual Redundancy Algorithm}
\label{sec:appendix_algo}
In~\cref{sec:method_tools}, we propose to minimize the visual redundancy to select frames. Solving~\cref{eq:dpp} is NP-hard, we use fast algorithm introduced in~\cite{fast_dpp} to solve~\cref{eq:dpp}. The algorithm only includes a small number of inner product of two
vectors, and can run efficiently. The complete algorithm is summarized in~\cref{alg:fast_dpp}. To get $K$ images from $T$ images,~\cref{alg:fast_dpp} runs in $\mathcal{O}(K^2T)$ and needs additional space $\mathcal{O}(KT)$. The stopping criteria is $\lvert\mathcal{S}\rvert=K$. We introduce a small number $\epsilon=1e^{-5}$ and add $d_j^2<\epsilon$ to the \textit{stopping criteria} for numerical stability of calculating $1/d_j$ following~\cite{fast_dpp}.

\section{Evaluation Details}
\label{sec:appendix_eval}
\noindent\textbf{Context length and multi-turn thinking}. Adapting VLMs for video understanding tasks often suffer from limited context length. In our experiments, if the token length exceeds the model limit, we discard earlier rounds until the model length limit is met. Another approach to reduce context length is only using single-turn thinking, \ie only allow models use at most one tool to solve problem. We compare this design with only single-turn mode in~\cref{tab:appendix_exp_turn}. It is shown that multi-turn mode gets significant gains compared with single-turn mode. So we use multi-turn mode that allows VLMs to think and call tools in multiple rounds.

\noindent\textbf{Decoding parameters}. For \ours-7B and Qwen2.5-VL-7B, we use temperature=0, top-k=1, top-p=0.001, following the Qwen official repository. For GPT-4o, we set temperature to 0 with the OpenAI Python SDK.

\section{More Experimental Results}
\label{sec:appenidx_results}
\noindent\textbf{Ablation on fine-tuned data}. To show the effectiveness of data with visual thinking trajectories, we run ablation on different data with \ours-7B. Results are listed in~\cref{tab:ablation_data} and show that training data with visual thinking trajectories will significantly improve the performance of the base model.

\noindent\textbf{Results on approach for frame selection}. First, we compare our method with recently proposed training-free methods, such as such as Q-Frame~\cite{q_frame} and AKS~\cite{AKS}. The results are shown in~\cref{tab:comp_frame_select}, which shows that our method achieves the best performance on EgoSchema~\cite{egoschema}. Second, we run ablation with different models for frame selection as shown in~\cref{tab:ablation_clip}. It is noticed that SigLIP-2~\cite{tschannen2025siglip2} achieves the best result and we choose SigLIP-2 for frame selection in all experiments.

\begin{figure}
    \centering
    \includegraphics[width=0.99\linewidth]{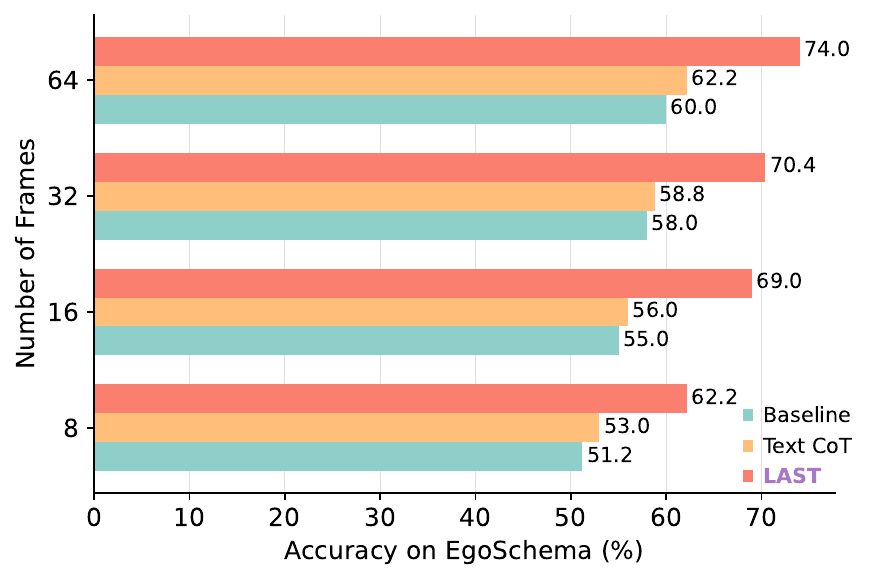}
    \vspace{-0.1in}
    \caption{Performance on EgoSchema under different frames of Qwen2.5-VL-7B (baseline), text CoT and \ours-7B (ours).}
    \label{fig:frames}
\end{figure}

\noindent\textbf{Results on different frames}. ~\cref{fig:frames} shows results of different thinking mode under different visual context length, \ie different frames. It is shown that \ours still outperforms text-only CoT~\cite{cot} under different frames. 

\begin{figure}
    \centering
    \includegraphics[width=0.95\linewidth]{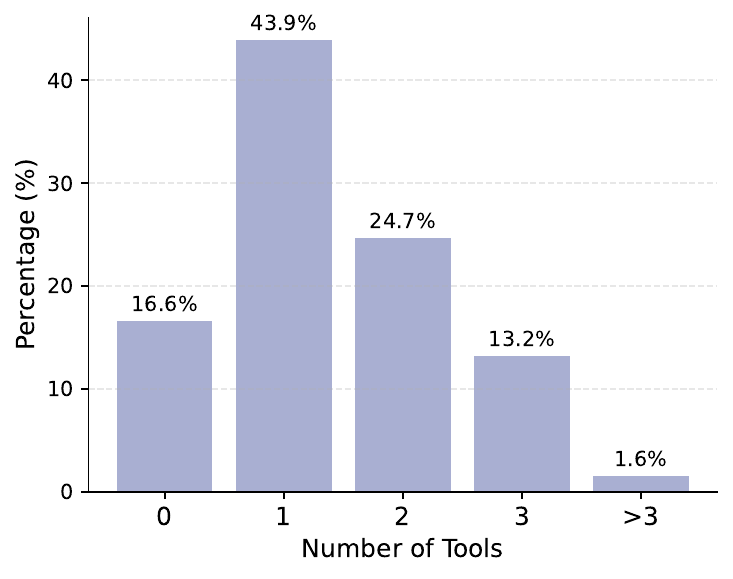}
    \caption{Distribution of number of tools used in VSI-Bench with GPT-4o.}
    \label{fig:num_tools}
\end{figure}

\noindent\textbf{The number of tools usage}. We count the number of tools used for each question in VSI-Bench~\cite{vsibench} with GPT-4o~\cite{gpt4o}, as shown in~\cref{fig:num_tools}. We observe that most questions use one or two tools to be solved, which accounts for over half of the questions.

\noindent\textbf{Results on image benchmarks}. ~\cref{tab:image} shows results on three image benchmarks. \ours shows significant gains in image benchmarks, indicating that \ours can also be applied to 2D image scenarios.

\noindent\textbf{More models}.  We further evaluate \ours for Gemini-2.5-flash~\cite{comanici2025gemini2.5} and results are shown in~\cref{tab:appendix_other_models}. \ours consistently improves the performance of both models, which demonstrates the generalizability of our method.

\noindent\textbf{Cost analysis}. The token cost of three tasks is presented in~\cref{tab:appendix_token}. From~\cref{tab:appendix_token}, we can see that \ours uses more visual tokens, while text-only CoT uses more text tokens. Although \ours requires more tokens than the base model with text CoT, the increased latency is not significant. This is because visual tokens are usually input in the \textit{prefill} stage, which is compute-bound and can be \textit{highly parallelized}. In our settings, we deploy \ours-7B on a single A800 GPU with 32K context with vLLM. The throughput of \ours-7B achieves $>$9000 tokens per second. In addition, for lightweight tool models, we also deploy each model on a single A800 GPU with Triton Inference Server. For all tasks, the average increased latency compared with using Qwen2.5-VL-7B~\cite{bai2025qwen25vl} is less than 1 second. 

\section{Additional Qualitative Results}
\begin{figure*}
    \centering
    \includegraphics[width=0.95\textwidth]{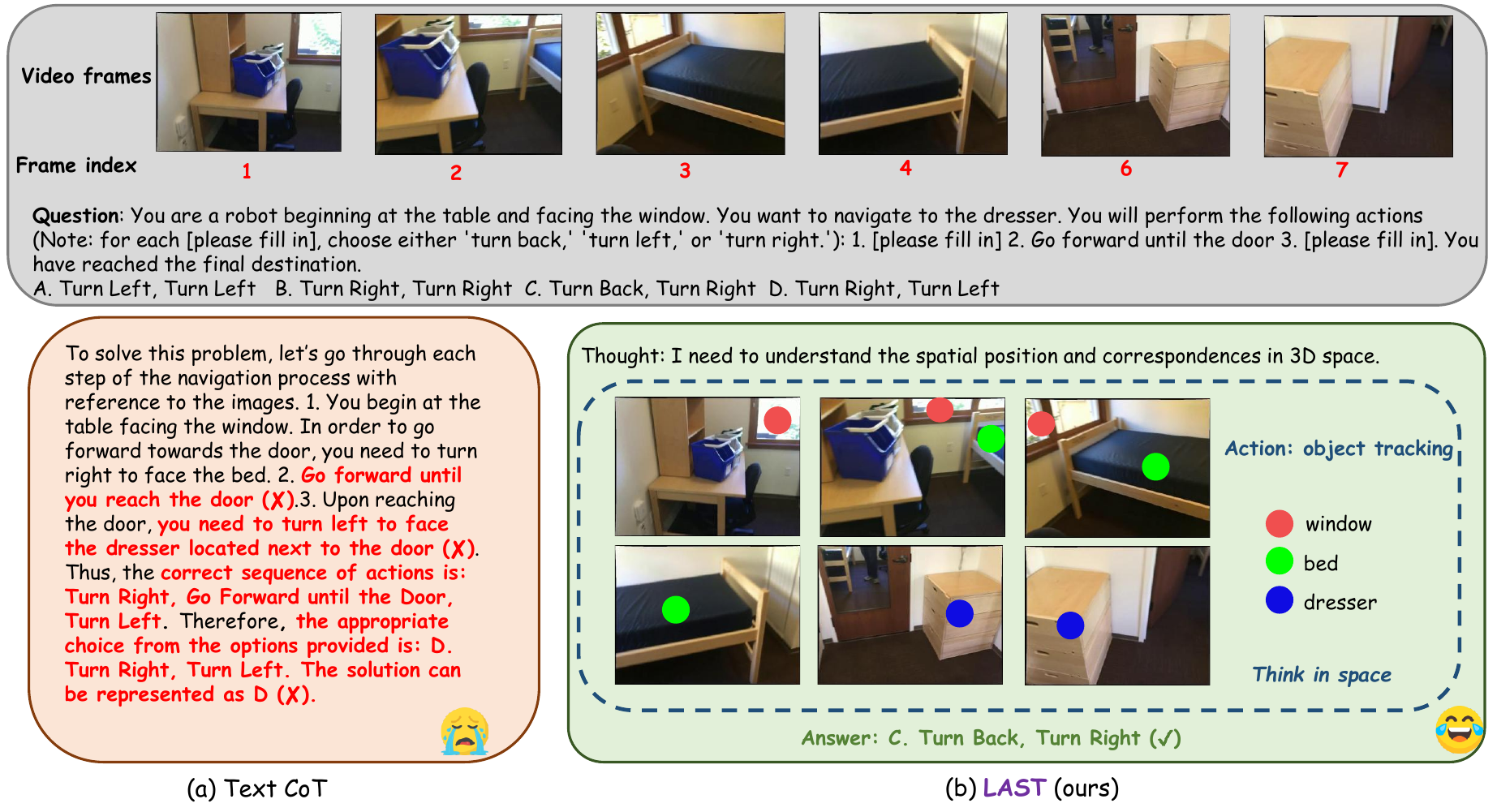}
    \caption{Qualitative comparison of text-only CoT and \ours on VSI-Bench~\cite{vsibench}. We highlight \textcolor{red}{errors} in text-only CoT and present an error analysis. With text CoT, GPT-4o fails to understand object correspondce in 3D space.}
    \label{fig:example2}
\end{figure*}

\begin{figure*}
    \centering
    \includegraphics[width=0.95\textwidth]{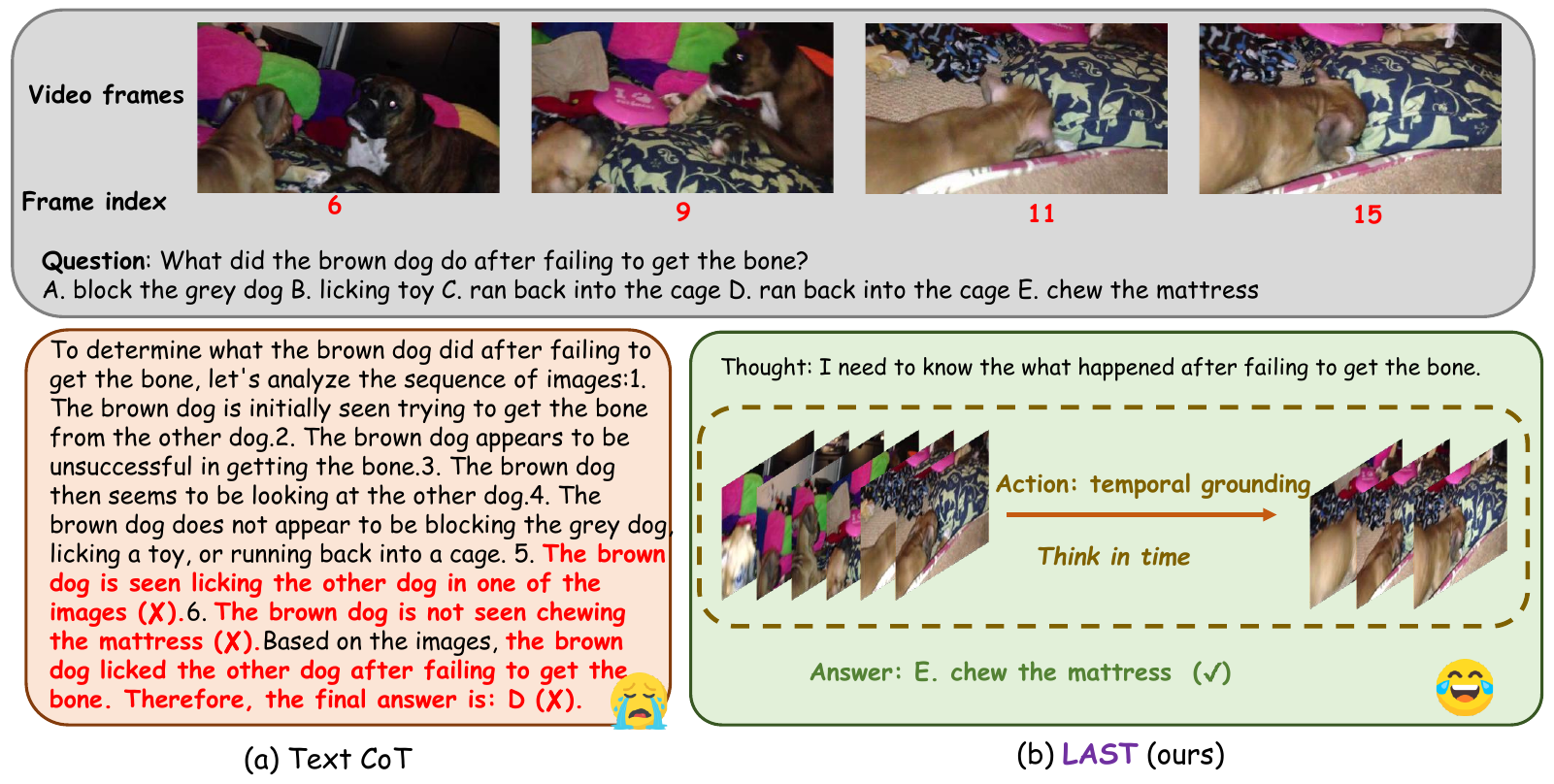}
    \caption{Qualitative comparison of text-only CoT and \ours on NExT-QA~\cite{nextqa}.}
    \label{fig:example3}
\end{figure*}

We show more qualitative results on VSI-Bench~\cite{vsibench} in~\cref{fig:example2} and NExT-QA~\cite{nextqa} in~\cref{fig:example3}.

\begin{table}[t]
\centering
\caption{A detailed breakdown of the datasets for fine-tuning.}
\vspace{-0.1in}
\begin{tabular}{lccc}
    \toprule
    \textbf{Type} & \textbf{Dataset} & \textbf{Split} & \textbf{Size} \\
    \midrule
    \multirow{1}{*}{Video} 
        & LLaVA-Video~\cite{zhang2025llavavideo} & -- & 178K \\
    \midrule
    \multirow{2}{*}{3D Scene} 
        & ScanQA~\cite{azuma2022scanqa} & Train & 25K \\
        & SQA3D~\cite{sqa3d} & Train & 16K \\
    \midrule
    \multirow{2}{*}{Image} 
        & CogCom~\cite{qi2025cogcom} & -- & 41K \\
        & DeepEyes~\cite{deepeyes} & -- & 47K \\
    \bottomrule
\end{tabular}
\label{tab:data}
\end{table}
\begin{algorithm}[t]
\caption{Fast Greedy MAP Inference}
\begin{algorithmic}[1]
\State \textbf{Input:} Kernel $L_{p,q}$, \textit{stopping criteria}
\State \textbf{Initialize:} $c_i = []$, $d_i^2 = L_{ii}$, $j = \arg\max_{i \in \mathcal{I}} \log(d_i^2)$, $\mathcal{S} = \{j\}$
\While{\textit{stopping criteria} not satisfied}
    \For{$i \in \mathcal{I} \setminus \mathcal{S}$}
        \State $e_i = \left(L_{ji} - \langle c_j, c_i \rangle\right) / d_j$
        \State $c_i \leftarrow [c_i ~~ e_i]$
        \State $d_i^2 \leftarrow d_i^2 - e_i^2$
    \EndFor
    \State $j = \arg\max_{i \in \mathcal{I} \setminus S} \log(d_i^2)$
    \State $S \leftarrow S \cup \{j\}$
\EndWhile
\State \textbf{Return:} $\mathcal{S}$
\end{algorithmic}
\label{alg:fast_dpp}
\end{algorithm}

\begin{table}
\centering
\caption{Comparison of single-turn and multi-turn mode.}
\vspace{-0.1in}
\begin{tabular}{lcccc}
    \toprule
    \textbf{Method} & \textbf{VSI-Bench} & \textbf{EgoSchema} \\
    \midrule
    GPT-4o & 40.9 & 69.6 \\
    + \ours (\textit{single-turn}) & 45.8 & 83.1 \\
    \rowcolor{myrowcolor}
    + \ours (\textit{multi-turn}) & \textbf{46.6} & \textbf{85.4}\\    
    \bottomrule
    \end{tabular}
\label{tab:appendix_exp_turn}
\end{table}
\begin{table}
\centering
\caption{\textbf{Ablation on data}. \texttt{T} denotes text CoT data and \texttt{V} denotes visual thinking data.}
\vspace{-0.1in}
\begin{tabular}{lccc}
    \toprule
     \textbf{Method} & \textbf{Data} & \textbf{VSI-Bench} & \textbf{EgoSchema} \\
     \midrule
     Qwen2.5-VL-7B & -- & 33.0 & 55.0 \\
     \ours-7B & \texttt{T} & 35.1 & 56.0\\
     \ours-7B & \texttt{V} & 40.2 & 68.2 \\
     \rowcolor{myrowcolor}
     \ours-7B & \texttt{T}+\texttt{V} & 41.3 & 69.0\\
    \bottomrule
    \end{tabular}
\label{tab:ablation_data}
\end{table}
\begin{table}[t]
\centering
\caption{Comparison of different method for frame selection. We compare different methods on EgoSchema~\cite{egoschema} with GPT-4o~\cite{gpt4o}. The number of input frames is set to 8.}
\vspace{-0.1in}
\resizebox{\linewidth}{!}{
    \begin{tabular}{lccc}
    \toprule
    \textbf{Method} & \textbf{Relevance} & \textbf{Redundancy} & \textbf{EgoSchema} \\
    \midrule
    Uniform sample & \no & \no & 69.6 \\
    Q-Frame~\cite{q_frame} & \yes & \no & 71.2 (\textcolor{mygreen}{+1.6})\\
    AKS~\cite{AKS}   & \yes & \no & 71.8 (\textcolor{mygreen}{+2.2})\\
    Ours & \yes & \yes & \textbf{73.6 (\textcolor{mygreen}{+4.0})}\\
    
    \bottomrule
    \end{tabular}
}
\label{tab:comp_frame_select}
\end{table}
\begin{table}
\centering
\caption{Ablation on models for frame selection. We use GPT-4o for EgoSchema and the number of frames is set to 8.}
\vspace{-0.1in}
    \begin{tabular}{lccc}
    \toprule
    \textbf{Model} & \textbf{Arch}. & \textbf{Res}. & \textbf{EgoSchema} \\
    \midrule
    CLIP~\cite{clip} & ViT-B/16 & 224$^2$ & 72.8\\
    EVA-CLIP~\cite{sun2023evaclip} & ViT-B/16 & 224$^2$ & 73.4\\
    Open-CLIP~\cite{openclip} & ViT-B/16 & 224$^2$ &  73.2\\
    SigLIP-2~\cite{tschannen2025siglip2} & ViT-B/16 & 224$^2$ & \textbf{73.6} \\
    \bottomrule
    \end{tabular}
\label{tab:ablation_clip}
\end{table}

\begin{table}[t]
\centering
\caption{\textbf{Evaluation results on three image benchmarks.}}
\vspace{-0.1in}
    \begin{tabular}{lcccc}
    \toprule
    \multirow{2}{*}{\textbf{Method}} & \multicolumn{2}{c}{\textbf{HR-Bench}} & \textbf{V}* & \textbf{BLINK} \\
    \cmidrule(lr){2-3} \cmidrule(lr){4-4} \cmidrule(lr){5-5}
    & 4K & 8K & -- & Val \\
    \midrule
    InternVL3-8B~\cite{zhu2025internvl3} & 70.0 & 69.3 & 70.2 & 55.5\\
    Qwen2.5-VL-7B~\cite{bai2025qwen25vl} & 68.8 & 65.3 & 76.4 & 56.4\\
    \rowcolor{myrowcolor}
    \ours-7B & 75.0 & 72.0 & 81.2 & 61.8 \\
    \midrule
    GPT-4o~\cite{gpt4o} & 65.0 &59.6 & 67.5& 63.3 \\
    \rowcolor{myrowcolor}
    + \ours & 69.0 & 63.0 & 71.6 & 64.2 \\ 
    \bottomrule
    \end{tabular}
\label{tab:image}
\end{table}
\begin{table*}
\centering
\caption{\textbf{Comparison on four benchmarks for Gemini-2.5-flash~\cite{comanici2025gemini2.5}}. We sample 16 frames for all benchmarks, except 8 frames for EgoSchema~\cite{egoschema}.}
\begin{tabular}{lccccc}
    \toprule
    \multirow{2}{*}{\textbf{Method}} & \multirow{2}{*}{\textbf{EgoSchema}} & \multirow{2}{*}{\textbf{LVBench}} & \multirow{2}{*}{\textbf{VSI-Bench}} & \multicolumn{2}{c}{\textbf{NExT-QA}}\\
    \cmidrule(lr){5-6}    
     & & & & Val & ATP-hard \\
     \midrule
    Gemini-2.5-flash~\cite{comanici2025gemini2.5} & 67.2 & 42.1 & 47.5 & 80.2 & 75.0  \\
    \rowcolor{myrowcolor}
    + \ours & 78.2 & 51.4 & 55.1 & 84.7 & 78.6\\  
    \bottomrule
    \end{tabular}
\label{tab:appendix_other_models}
\end{table*}
\begin{table*}
    \centering
    \caption{\textbf{Token cost} comparison on three benchmarks. We present average token usage per sample and performance of each benchmark for comparison.}
    \vspace{-0.1in}
    \resizebox{\textwidth}{!}
    {
        \begin{tabular}{l *{12}{r}}
        \toprule
         \multirow{2}{*}{\textbf{Method}} & \multicolumn{4}{c}{\textbf{VSI-Bench}} & \multicolumn{4}{c}{\textbf{NExT-QA}} & \multicolumn{4}{c}{\textbf{EgoSchema}} \\
        \cmidrule(lr){2-5}  \cmidrule(lr){6-9} \cmidrule(lr){10-13}
          & \# in. te. &  \# in. vis. & \# out. te. & perf.  & \# in. te. &  \# in. vis. & \# out. te. & perf. & \# in. te. &  \# in. vis. & \# out. te. & perf. \\
        \midrule
        Qwen2.5-VL-7B~\cite{bai2025qwen25vl} & 53.8  & 3128.0 & 1.7   & 33.0 & 54.9  & 2712.3 & 1.0   & 79.6 & 169.5 & 942.8  & 1.0   & 55.0 \\
        + CoT~\cite{cot}                     & 57.3  & 3128.0 & 276.2 & 35.2 & 57.9  & 2712.3 & 163.9 & 78.4 & 172.5 & 942.8  & 171.7 & 56.0 \\
        \ours-7B                             & 543.4 & 7384.1 & 78.1  & 41.3 & 543.9 & 4138.2 & 53.4  & 86.2 & 658.5 & 2108.4 & 102.8 & 69.0 \\        
        \bottomrule
        \end{tabular}    
    }
    \flushleft
    \vspace{-0.1in}
    \footnotesize {\textbf{Abbreviation}. in. te.: input text token, in. vis.: input visual token, out. te.: output text token, perf.: performance.}
\label{tab:appendix_token}
\end{table*}

\end{document}